\documentclass[lettersize,journal]{IEEEtran}
\usepackage{amsmath,amsfonts}
\usepackage{algorithmic}
\usepackage{algorithm}
\usepackage{array}
\usepackage[caption=false,font=normalsize,labelfont=sf,textfont=sf]{subfig}
\usepackage{textcomp}
\usepackage{stfloats}
\usepackage{url}
\usepackage{verbatim}
\usepackage{graphicx}
\usepackage{cite}
\usepackage[colorlinks,linkcolor=red,anchorcolor=blue,citecolor=green]{hyperref}	
\usepackage[switch]{lineno} 
\usepackage{multicol}
\usepackage{multirow}
\usepackage[export]{adjustbox}
\usepackage[linewidth=1pt]{mdframed}
\usepackage{gensymb}
\usepackage{threeparttable}
\usepackage{booktabs}
\usepackage{makecell}
\usepackage{titlesec}
\usepackage{xr}
\titlespacing\section{0pt}{12pt plus 4pt minus 2pt}{0pt plus 2pt minus 2pt}
\titlespacing\subsection{0pt}{12pt plus 4pt minus 2pt}{0pt plus 2pt minus 2pt}
\hypersetup{hidelinks}

\begin{document}
\switchlinenumbers	

\title{A Powered Prosthetic Hand with Vision System for Enhancing the Anthropopathic Grasp}

\author{Yansong Xu, Xiaohui Wang, Junlin Li, Xiaoqian Zhang, Feng Li, Qing Gao, Chenglong Fu, and Yuquan Leng*
\thanks{This work was supported in part by the National Natural Science Foundation of China under Grants 52175272, U1913205 and 62006204, in part by the Science, Technology, and Innovation Commission of Shenzhen Municipality under Grants JCYJ20220530114809021 and ZDSYS20200811143601004. \emph{(Corresponding author: Yuquan Leng)}}

\thanks{Yansong Xu, Xiaohui Wang, Xiaoqian Zhang and Feng Li are with the State Key Laboratory of Robotics, Shenyang Institute of Automation, Chinese Academy of Sciences, Shenyang 110016, China, and also with Institutes for Robotics and Intelligent Manufacturing, Chinese Academy of Sciences, Shenyang 110169, China, and also with University of Chinese Academy of Sciences, Beijing 100049, China (e-mail: xuyansong21@mails.ucas.ac.cn; wangxiaohui1@sia.cn; zhangxiaoqian21@mails.ucas.ac.cn; lifeng212@mails.ucas.ac.cn).}

\thanks{Junlin Li is with the State Key Laboratory of Robotics, Shenyang Institute of Automation, Chinese Academy of Sciences, Shenyang 110016, China, and also with the Institutes for Robotics and Intelligent Manufacturing, Chinese Academy of Sciences, Shenyang 110169, China (email: lijunlin@sia.cn).}

\thanks{Qing Gao is with School of Electronics and Communication Engineering, Sun Yat-sen University, Shenzhen 518107, China (e-mail: gaoqing.ieee@gmail.com).}

\thanks{Chenglong Fu and Yuquan Leng are with Department of Mechanical and Energy Engineering, Southern University of Science and Technology, Shenzhen 518055, China (e-mail: fucl@sustech.edu.cn; lengyq@sustech.edu.cn).}}



\maketitle

\begin{abstract}
	The anthropomorphism of grasping process significantly benefits the experience and grasping efficiency of prosthetic hand wearers. Currently, prosthetic hands controlled by signals such as brain-computer interfaces (BCI) and electromyography (EMG) face difficulties in precisely recognizing the amputees' grasping gestures and executing anthropomorphic grasp processes. Although prosthetic hands equipped with vision systems enables the objects' feature recognition, they lack perception of human grasping intention. Therefore, this paper explores the estimation of grasping gestures solely through visual data to accomplish anthropopathic grasping control and the determination of grasping intention within a multi-object environment. To address this, we propose the Spatial Geometry-based Gesture Mapping (SG-GM) method, which constructs gesture functions based on the geometric features of the human hand grasping processes. It's subsequently implemented on the prosthetic hand. Furthermore, we propose the Motion Trajectory Regression-based Grasping Intent Estimation (MTR-GIE) algorithm. This algorithm predicts pre-grasping object utilizing regression prediction and prior spatial segmentation estimation derived from the prosthetic hand's position and trajectory. The experiments were conducted to grasp 8 common daily objects including cup, fork, etc. The experimental results presented a similarity coefficient $R^{2}$ of grasping process of 0.911, a Root Mean Squared Error ($RMSE$) of 2.47\degree, a success rate of grasping of 95.43$\%$, and an average duration of grasping process of 3.07$\pm$0.41 s. Furthermore, grasping experiments in a multi-object environment were conducted. The average accuracy of intent estimation reached 94.35$\%$. Our methodologies offer a groundbreaking approach to enhance the prosthetic hand's functionality and provides valuable insights for future research.
\end{abstract}

\begin{IEEEkeywords}
	Computer vision, gesture modeling, full-automatic control, prosthetic hand system.
\end{IEEEkeywords}


\section{Introduction}
\IEEEPARstart{H}{ealthy} people mainly grasp and manipulate daily objects with their hands, while hand amputees lose the ability to do so. The absence of physical limbs brings extremely negative impact for the hand amputees. The powered prosthetic hand with fingers grasping objects as naturally as healthy people operate could help hand amputees regain grasping ability and self-esteem in daily life.

The motion control of prosthetic hands relies on the motion intent of the amputees. Currently, common techniques for controlling prosthetic hands using human intent signals include brain-computer interfaces (BCI) \cite{9302890,aggarwal2022review} and electromyography (EMG) \cite{malevsevic2018vector,xiong2022learning}. The motion intent of the human hand originates from the brain. Amputees control prosthetic hands through the extraction of electroencephalography (EEG) signals from the cerebral cortex using BCI, enabling motion intent recognition \cite{zhang2020novel,li2018approach}. Although invasive BCI offers superior signal quality compared to non-invasive BCI, the invasive approach requires the implantation of microsensors into the cerebral cortex, often causing harm to the human body. On the other hand, EEG, as a non-invasive technique, involves placing electrodes on the scalp without penetrating the skin or skull, thus causing no adverse effects on human tissues. However, EEG signals suffer from poor signal-to-noise ratio, susceptibility to environmental influences, and low intent resolution \cite{jeong20222020,wang2020common}. Therefore, current BCI-based methods face challenges in achieving fine-grained motion control of prosthetic hands.

Using residual limb EMG signals for controlling prosthetic hands is the most commonly used method. The earliest related studies focused on simple actions of prosthetic hand opening and closing using EMG thresholds \cite{geethanjali2016myoelectric}. Subsequently, researchers proposed a series of pattern recognition methods for EMG signals to capture a wider range of human gestures. For instance, Malesevic et al. \cite{malevsevic2018vector} analyzed motion features of fingers with multi-channel EMG signal via a vector autoregressive hidden Markov model. Bhagwat et al. \cite{bhagwat2020electromyogram} classified finger movements by sparse filtering with wavelet packet coefficients. Mendes et al. \cite{mendes2022surface, li2024improving} recognized gestures with a convolutional neural network (CNN). These methods primarily focused on gesture classification rather than continuous gesture signal recognition. Furthermore, using residual limb EMG signals for prosthetic hand control faces challenges such as large inter-individual variations in EMG signals, muscle atrophy of amputees, signal interference caused by sweat, and electrode displacement \cite{liu2016development, 10436655}. Therefore, achieving fine-grained motion control of prosthetic hands based on EMG signals remains challenging at present.

When humans grasp objects, they adjust their grasping gestures based on shape, size, and other characteristics of the objects. Similarly, the control of prosthetic hands also relies on the feature information of the objects to perform adaptive grasping gestures \cite{sun2016object}. Therefore, researchers have proposed prosthetic hand systems equipped with vision sensors. Currently, most prosthetic hands with vision sensors only obtain relevant information about the pre-grasping object (such as its category, shape, and size) and estimate human motion intent by incorporating other sensor information. For example, Markovic et al. \cite{markovic2015sensor,starke2022semi} tracked the arm's motion state with inertial measurement unit (IMU), Castro et al. \cite{castro2022continuous,starke2022semi} measured the relative distance between the hand and the object with distance sensors, and Karrenbach et al. \cite{karrenbach2022improving,zandigohar2024multimodal, shi2023gsi} monitored eye gaze direction with eye trackers. Although the addition of sensors provides more perceptual information to the system, it also increases the cost and complexity of the system. If the vision sensor and the captured data help obtain the information of the pre-grasping object and the motion intent of the human, the complexity of system can be decreased while ensuring functionality. The general control procedure of vision-based prosthetic hand systems is as follows: First, several specific grasping gestures for different objects are predefined within the system. Subsequently, during the grasping process, the system perceives and analyzes the characteristic attributes of the object. Ultimately, the system selects the appropriate gesture to accomplish the grasping task. However, the variation of gestures during the transition from pre-grasping to final grasping has not been fully considered in prosthetic hand control. If the prosthetic hand can exhibit anthropomorphic gestures during this process, it would significantly help hand amputees regain self-esteem.

To address the above-mentioned challenges, this paper aims to investigate the control of prosthetic hand with vision that can perform human-like and natural grasping gestures. Due to the lack of direct perception of biological signals for controlling hand movements, achieving anthropomorphic gesture control during the grasping process and estimating grasping intent in a multi-object environment solely based on visual information confronts with significant challenges. In this paper, the spatial variation regularity of hand gestures during the processes of grasping objects were explored. Thus, an innovative Spatial Geometry-based Gesture Mapping (SG-GM) method was proposed. This method incorporates spatial geometry, including geometric features of the target object, hand-object distance and grasping gesture. By utilizing gesture data from healthy hands during object grasping, the SG-GM method constructs gesture transformation functions based on the hand-object distance for different objects. These functions are then deployed on prosthetic hand with vision sensor to achieve anthropomorphic grasping gestures for the wearer. Additionally, the Motion Trajectory Regression-based Grasping Intent Estimation (MTR-GIE) algorithm was proposed for estimating the pre-grasping objects in a multi-object environment for the prosthetic hand. Finally, experiments were conducted by grasping eight common daily objects including cup, fork, etc. Experiments of grasping in multi-object environment were also conducted to verify the effectiveness of the proposed research methodology.

The contributions of the paper are listed as follows: 
\begin{enumerate}
	\item{This paper represents the first implementation of anthropopathic grasping gestures for the prosthetic hand. Previous researches focused more on the grasping result while taking less consideration of the gesture variations during grasping process. Anthropomorphic grasping gestures help hand amputees regain self-esteem and improve grasping efficiency.}
	\item{The Spatial Geometry-based Gesture Mapping (SG-GM) method was proposed. This method enables the construction of gesture transformation functions for different object-grasping processes based on grasping gesture data from human hands. These functions can then be deployed on prosthetic hand with vision sensor to perform anthropomorphic grasping process.}
	\item{The Motion Trajectory Regression-based Grasping Intent Estimation (MTR-GIE) algorithm was proposed. This algorithm allows the prosthetic hand system to estimate the grasping intent in multi-object environment based on vision sensor solely.}
\end{enumerate}

These contributions collectively improve the performance and availability of the prosthetic hand in grasping tasks. This paper is organized as following sections. Section II presents an overview of the system. In section III, we conduct grasping experiments of the prosthetic hand system. Section IV presents the experimental results and analysis. The discussion is presented in section V, and conclusion is stated in section VI.
\vspace{-2pt}
\begin{figure*}[!t]
	\vspace{-10pt}
	\centering
	\includegraphics[width=\textwidth]{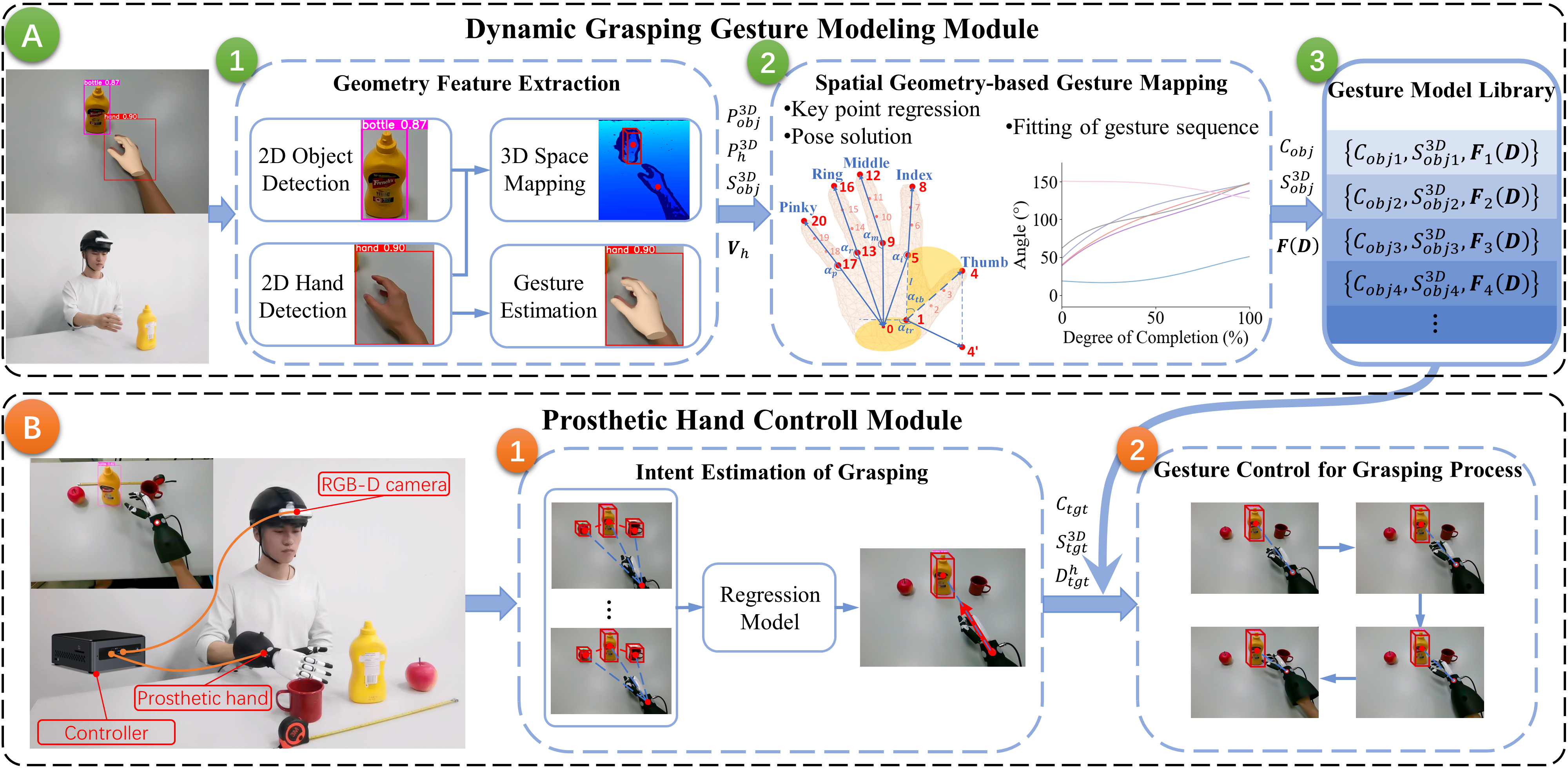}
	\caption{Overview of the intelligent prosthetic hand system. (a) Diagram A represents the workflow chart of the dynamic grasping gesture modeling module. (b) Diagram B represents the workflow chart of the prosthetic hand control module. The dynamic grasping gesture modeling module first extracts the geometric features of the hand and the objects as presented in sub-diagram A1. Then, it maps the hand posture angles into spatiotemporal gesture transformation functions as presented in sub-diagram A2. Finally, it stores the object properties and gesture functions into the gesture model library as presented in sub-diagram A3. The prosthetic hand control module starts by quickly identifying the target object using an intent estimation algorithm as presented in sub-diagram B1. It then retrieves the corresponding gesture function from the gesture model library in sub-diagram A3. Based on the information in the gesture function, the prosthetic hand is guided to perform natural grasping gestures during the movement as presented in sub-diagram B2.}
	\label{fig1}
	\vspace{-0.3cm}
\end{figure*}

\vspace{-2pt}
\section{System and Method}
The hardware of the prosthetic hand system with vision sensor is composed of three parts (as presented in Fig. \ref{fig1}): an RGB-D camera (Intel RealSense D455) worn on the user's head, a controller (Ubuntu 20.04 LTS, CPU 12th Gen Intel\textsuperscript{\textregistered} Core™ i7-12700 2.10 GHz, GPU NVIDIA GeForce RTX 3060 Ti), and a five-finger dexterous hand (Inspire RH56BF3-2R). The size of the dexterous hand is comparable to that of a human hand, with six degrees of freedom (DOFs).

As presented in Fig. \ref{fig1}, the system consists of two sub-functional modules: a dynamic grasping gesture modeling module and a prosthetic hand control module. The function of the dynamic grasping gesture modeling module utilizes the SG-GM method to extract the geometry features of human hand grasping objects, construct gesture transformation functions for different objects' grasping processes, and generate a gesture model library. The prosthetic hand control module consists of pre-grasping object intent estimation and grasping control. On one hand, it achieves the estimation of pre-grasped objects by using the MTR-GIE algorithm for the prosthetic hand in a multi-object environment. On the other hand, it enables the anthropomorphic grasping of the prosthetic hand.

\subsection{Dynamic grasping gesture modeling module}
\begin{figure*}[!t]
	\vspace{-10pt}
	\centering
	\includegraphics[width=\textwidth]{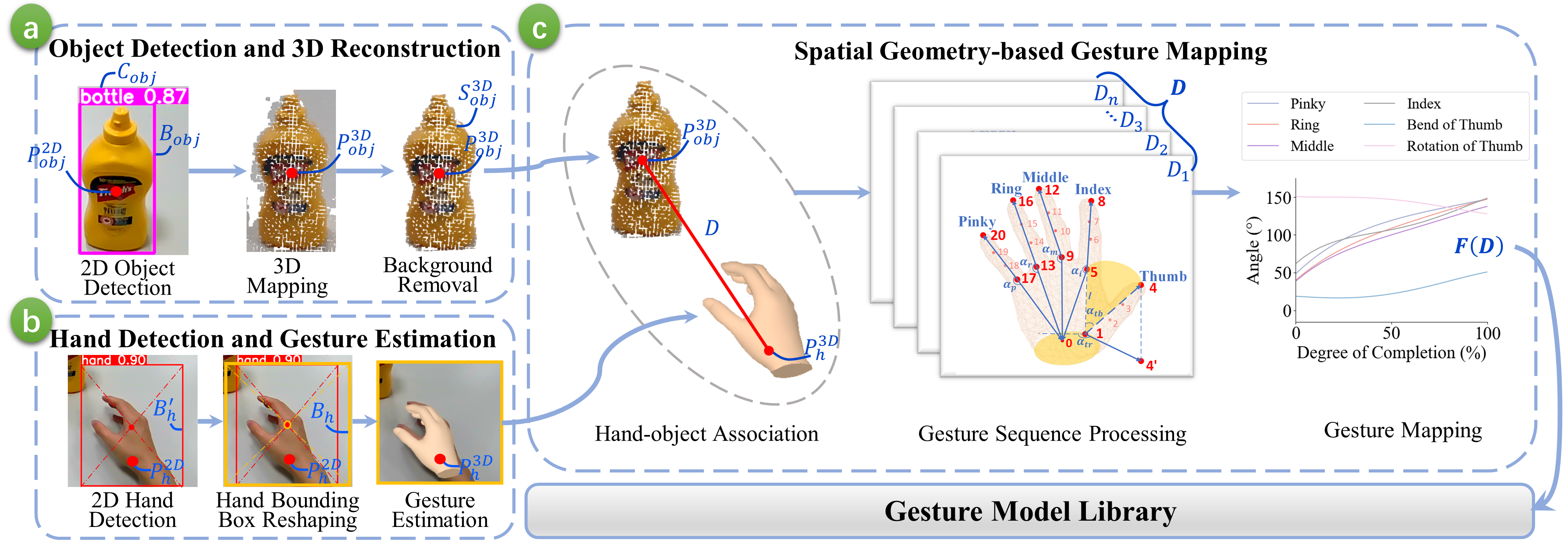}
	\caption{Pipeline of the SG-GM method. In the object detection and 3D reconstruction part as presented in sub-diagram A, the process consists of 2D object detection, 3D mapping, and background removal in order to obtain 3D model parameters and object positions. In the hand detection and gesture estimation part as presented in sub-diagram B, the steps involve 2D hand detection, reshaping the hand bounding box, and gesture estimation to obtain the gesture and hand positions for each frame. Subsequently, the information of the hand and the object is associated, and the gesture transformation functions that are spatiotemporally related to the object are derived as presented in sub-diagram C. The sequence $\boldsymbol{D}=\{D_1, D_2, D_3, \ldots, D_n\}$ represents the hand-object distance during the hand's movement. In the gesture mapping section, the horizontal axis represents the spatial distance progress of the hand from the starting moving position to the final grasping position.}
	\label{fig2}
	\vspace{-0.3cm}
\end{figure*}

This article takes the right prosthetic hand as the research example. The data for the dynamic grasping gesture modeling module are obtained from individuals grasping daily objects with healthy hands. The SG-GM method used in this module follows the process presented in Fig. \ref{fig2}, which includes object detection and 3D reconstruction, hand detection and gesture estimation, spatial geometry-based gesture mapping, and gesture model library construction.

In the process of dynamic gesture modeling, the camera coordinate system is defined as the world coordinate system when the hand is detected for the first time. In order to avoid cumulative errors during positioning, the world coordinate system is automatically reinitialized with each new grasping movement. Since the data for each grasp are independent from one another, and relative spatial relationships are adopted in the experiments at the same time, redefining the world coordinate system does not affect the gesture space mapping.

\subsubsection{Object detection and 3D reconstruction}
To obtain the necessary feature information of objects, three steps are performed: object detection, 3D mapping, and background removal.

2D object detection aims to obtain the position information of objects in the image. For object detection, the lightweight object detection network model YOLOv5 \cite{yolov5} is used. With the RGB image as the input to the model, the object's class $C_{obj}$ and bounding box $B_{obj}$ can be obtained. The center point of the bounding box is then used as the 2D position $P_{obj}^{2D}$ of the object.

For 3D mapping, it is necessary to compute the 3D coordinates $(x,y,z)$ in the world coordinate system for the corresponding points $(u,v)$ in the pixel coordinate system. The calculation was performed by using the depth information obtained from the depth image. In addition, the camera intrinsic parameters are required, which consist of the camera principal point $(p_x,p_y)$ and the focal lengths $(f_x,f_y)$ in the $x$ and $y$ directions, respectively. Specifically, for each point $(u,v)$ in the image, the depth value $h$ can be calculated. Then, within the region of bounding box $B_{obj}$, we apply the following equations to compute the 3D coordinates $(x,y,z)$:
\vspace*{-0.3\baselineskip}
\begin{equation}
	\label{eq1}
	\left\{\begin{array}{l}
		x=\frac{u-p_x}{f_x} \cdot z \\
		y=\frac{v-p_y}{f_y} \cdot z \\
		z=h.
	\end{array}\right.
	\vspace*{-0.3\baselineskip}
\end{equation}

A rough model of the 3D mapping of the region within bounding box $B_{obj}$ can be obtained via the mentioned equations. Furthermore, based on the 2D position $P_{obj}^{2D}$, the 3D position $P_{obj}^{3D}$ of the object in the world coordinate system can be determined.

The purpose of background removal is to eliminate the influence of outliers on the target 3D point cloud model. The background removal process consists of three steps. First, the Region of Interest (ROI) method is employed to capture the target detection box as the area of interest. Based on the location and size of the detection box $B_{obj}$, only the points within the detection box are retained in the entire point cloud. In the depth measurements, the bounding box contains background regions and potential outliers on the object, which need to be removed. Therefore, the second step involves depth threshold segmentation, where the object position $P_{obj}^{3D}$ is taken as the reference center, and twice the width of the target object is set as the depth threshold, thereby removing obvious background and foreground occlusions. Finally, K-means point cloud segmentation is applied. The point cloud of the target object exhibits significant inter-cluster differences from the background point cloud in 3D space. By applying K-means point cloud segmentation, a clean 3D point cloud model of the target object is generated, thus obtaining its size parameters $S_{obj}^{3D}$.

\subsubsection{Hand detection and gesture estimation}
To obtain information of hand gestures, three steps are performed: 2D hand detection, hand bounding box reshaping, and gesture estimation.

In the step of 2D hand detection, the detection of human hands is based on our improved SRHandNet detection model \cite{wang2019srhandnet}. We made adjustments to the SRHandNet heatmap regression of key points, increasing the threshold for missing key points from 5 to 15 and designating the wrist point as an indispensable key point. These adjustments ensure the robustness of the algorithm with severely occluded fingers. The model regresses the bounding box $B_{h}^{\prime}$ for the hand region in the RGB image and estimates the 2D position $P_{h}^{2D}$ of the wrist key point, as presented in Fig. \ref{fig2}.

In the step of hand bounding box reshaping. The original output bounding box $B_{h}^{\prime}$ from the SRHandNet model does not have a fixed aspect ratio, which can lead to distortion in the estimated gestures. To prevent the aspect ratio of the hand region image from bring negative impacts on the gesture estimation, it is necessary to correct the original bounding box $B_{h}^{\prime}$. We take the longest side of the original hand bounding box $B_{h}^{\prime}$ and construct a square bounding box with the same side length, ensuring that their centroids coincide. This results in the reshaped square bounding box $B_{h}$.

In the step of gesture estimation, the hand may appear at any position within the camera's field of view due to the relative motion between the hand and the camera during the grasping process. However, the original IntagHand gesture estimation model \cite{li2022interacting} is not capable of detecting hands, making it unable to locate and estimate gestures for small size of hand features in the entire image. To address this issue, we made improvements to the IntagHand gesture estimation method. By providing the information of the hand bounding box, the model is able to detect and localize the hand region. In this paper, based on the reshaped hand bounding box $B_h$, the hand region image is cropped with an aspect ratio of 1:1, and the cropped image is input into the IntagHand model, achieving undistorted gesture estimation at any position in the image. The improved IntagHand method enables us to obtain 778 hand mesh vertices $\boldsymbol{V}_{h}$, which are then used to construct the MANO hand model \cite{romero2022embodied,hasson2019learning}. The rendered hand model during hand-object grasping is presented in Fig. \ref{fig2}. Additionally, $P_{h}^{2D}$ can be mapped to obtain the 3D position $P_{h}^{3D}$ of the wrist key point via equations (\ref{eq1}).

\subsubsection{Spatial geometry-based gesture mapping}
In order to achieve smooth and effective control of the prosthetic hand based on the estimated gestures, it is necessary to perform hand-object association, gesture sequence processing, and gesture mapping on the data obtained during the grasping process.

In the step of hand-object association, the spatial geometric information of the hand and the object is highly correlated. This includes geometric features of the pre-grasping object, hand-object distance, and grasping gesture. In this paper, the related information was utilized to establish the gesture mapping relationship. First, in order to measure the spatial position relationship between the object and the hand, the Euclidean distance sequence between the object and the hand in the grasping process was defined as $\boldsymbol{D}$, as presented in the following equation.
\vspace*{-0.3\baselineskip}
\begin{equation}
	\label{eq2}
	\boldsymbol{D}=\left\|\boldsymbol{P}_h^{3 D}-\boldsymbol{P}_{o b j}^{3 D}\right\|_2.
	\vspace*{-0.3\baselineskip}
\end{equation}

This sequence $\boldsymbol{D}$ represents the distance changes between the hand and the object at different time stamps.

Gesture sequence processing comprises hand key points regression and pose estimation. 1) Hand key points regression: We adopted the joint regression method proposed by Loper et al. \cite{romero2022embodied,loper2015smpl} for the MANO hand model. However, the MANO hand model only provides the coordinates of 16 key hand joints through regression matrices. Therefore, the coordinates of 5 fingertip nodes, corresponding to the fingertips of each finger, are additionally extracted from the matrices of 778 vertices (with a size of 778 $\times$ 3) at specified positions [744, 320, 443, 554, 671]. The node positions are presented in Fig. \ref{fig2}, where the red marked points represent the locations of the corresponding nodes. 2) Pose estimation: Since the prosthetic hand is controlled based on the angles formed by the metacarpophalangeal joint-fingertip lines and a reference plane or axis, the gesture is mapped to an angle vector $\boldsymbol{\alpha}=\left[\alpha_p, \alpha_r, \alpha_m, \alpha_i, \alpha_{t b}, \alpha_{t r}\right]$. Here, $\alpha_p$, $\alpha_r$, $\alpha_m$, $\alpha_i$, $\alpha_{tb}$ represent the bending angles of the pinky finger, ring finger, middle finger, index finger, and thumb, respectively, while $\alpha_{tr}$ denotes the rotation angle of the thumb. The reference plane used for computing $\alpha_p$, $\alpha_r$, $\alpha_m$, $\alpha_i$, $\alpha_{tr}$ is the palm plane, which is fitted by the key points 0, 5, 9, 13, and 17. The reference axis for $\alpha_{tb}$ is the thumb rotation axis formed by key points 1 and 5.

In the step of gesture mapping, the gesture sequence during the grasping process is defined as $\boldsymbol{F}(\boldsymbol{D})=\left[F_p, F_r, F_m, F_i, F_{t b}, F_{t r}\right]$. Here, $F_p$, $F_r$, $F_m$, $F_i$, $F_tb$, $F_{tr}$ are polynomial fitting functions of the angle vector $\boldsymbol{alpha}$ with respect to the distance $\boldsymbol{D}$. An ideal gesture transformation curve can be obtained through the implementation of fourth-order polynomial fitting as the following equation:
\vspace*{-0.3\baselineskip}
\begin{equation}
	\label{eq3}
	F_j(\boldsymbol{D})=a_{4, j} \boldsymbol{D}^4+a_{3, j} \boldsymbol{D}^3+a_{2, j} \boldsymbol{D}^2+a_{1, j} \boldsymbol{D}^1+a_{0, j},
	\vspace*{-0.3\baselineskip}
\end{equation}where $a_{0, j}$, $a_{1, j}$, $a_{2, j}$, $a_{3, j}$, $a_{4, j}$ represent the coefficients of the polynomial, and $j \in\{p, r, m, i, t b, t r\}$. After regression and fitting, the obtained gesture function $\boldsymbol{F}(\boldsymbol{D})$ express a continuous and smooth gesture transformation process.

\subsubsection{Library construction of the gesture models}

The gesture model library is constructed based on the process of hand grasping objects. The data representation format of the model library is as follows:
\vspace*{-0.3\baselineskip}
\begin{equation}
	\label{eq4}
	\left\{\boldsymbol{C}_{obj}, \boldsymbol{S}_{obj}^{3D}, \boldsymbol{F}(\boldsymbol{D})\right\}.
	\vspace*{-0.3\baselineskip}
\end{equation}

The model library is designed to be scalable, allowing the construction of different gesture data based on the variation of the grasped objects and size parameters. Each entry in the library consists of the object category $\boldsymbol{C}_{obj}$, the 3D size information $\boldsymbol{S}_{obj}^{3D}$, and the functional representation $\boldsymbol{F}(\boldsymbol{D})$ of the grasping gesture sequence. The symbol $\left\{\boldsymbol{C}_{obj}, \boldsymbol{S}_{obj}^{3D}, \boldsymbol{F}(\boldsymbol{D})\right\}$ indicates that there are multiple entries in the model library, corresponding to different hand-object interactions and grasping actions.

\subsection{Prosthetic hand control module}
The control process of the prosthetic hand is presented in Fig. \ref{fig1} with the previously mentioned world coordinate system. The control method minimizes intervention in the subjective intent and behavior of users as much as possible. In the process of movement, the prosthetic hand system only controls the changes in gestures, while the position of the hand is controlled by the user. The control process is divided into three stages in chronological order: 1) The stage of intent estimation starts with detecting the prosthetic hand's position and ends with confirming the objects being grasped. The prosthetic hand maintains a steady semi-closed fist gesture within this stage. 2) The stage of grasping. Once the target object is determined, the prosthetic hand performs an anthropomorphic gesture. 3) The stage of grip tightening. After performing an anthropomorphic grasping gesture, the gesture will be fine-tuned in order to hold the object firmly.

\begin{figure}[!t]
	\vspace{-10pt}
	\centering
	\includegraphics[width=0.43\textwidth,keepaspectratio]{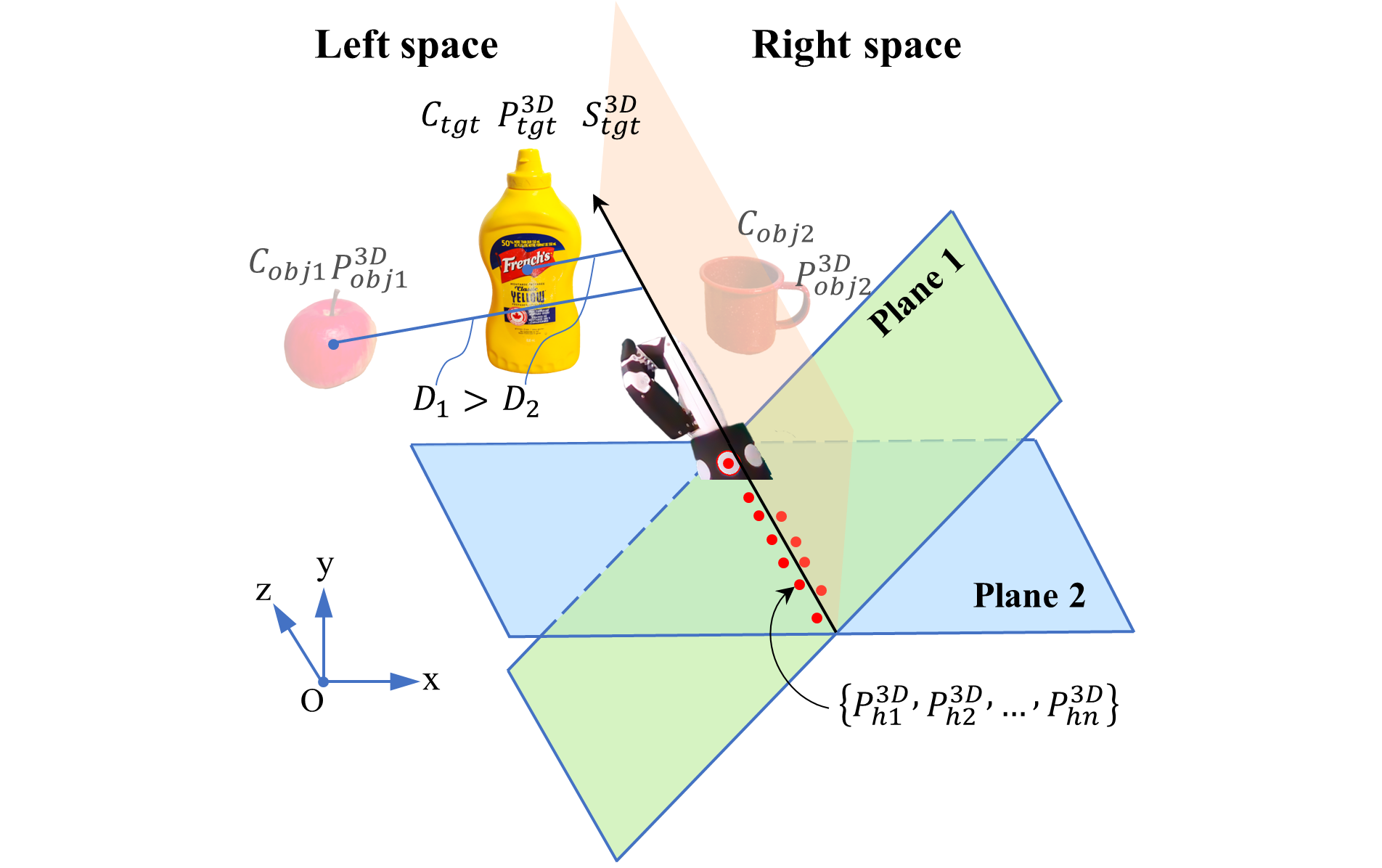}
	\caption{Diagram of the MTR-GIE algorithm for estimating grasping intent. The coordinate in the figure represents the world coordinate. The red points indicate the position of the wrist during the movement of the prosthetic hand. The MTR-GIE algorithm predicts the future movement trajectory based on the historical positions of the prosthetic hand during the grasping process. The intersection of the regression plane 1 marked in green and the regression plane 2 marked in blue in the figure represents the predicted regression line. This regression line is used as the predicted trajectory. The plane marked in orange on the predicted trajectory serves as a spatial separation plane to separate the objects in the left space (for right-hand prosthetic). The object closest to the left space is then estimated as the target object.}
	\label{fig3}
	\vspace{-0.3cm}
\end{figure}

\subsubsection{Intent estimation for grasping objects}

During the process of grasping objects, the motion trajectory of the wrist is approximated as a straight line in space. Based on this principle, the MTR-GIE algorithm is proposed in this paper. The positions of $n$ prosthetic hand wrists, represented as $\left\{P_{h 1}^{3 D}, P_{h 2}^{3 D}, \ldots, P_{h n}^{3 D}\right\}$, where $n\geq3$, are collected. By performing regression analysis on these wrist positions, a spatial straight line is obtained. This enables us to predict the direction of prosthetic hand movement and estimate the intent of grasping, as presented in Fig. \ref{fig3}. To enhance the localization of the prosthetic hand wrist, white markers are attached to it. In 3D Euclidean space, a straight line is obtained by the intersection of two planes. Consequently, two plane equations are formulated to represent this regression line.
\vspace*{-0.3\baselineskip}
\begin{equation}
	\label{eq5}
	\left\{\begin{array}{l}
		x=w_1 y+w_2 z+w_0 \\
		y=w_1^{\prime} x+w_2^{\prime} z+w_0^{\prime}.
	\end{array}\right.
	\vspace*{-0.3\baselineskip}
\end{equation}

For plane 1: $x=w_1 y+w_2 z+w_0$, the residual sum of squares $Loss$ is defined via the following equation: 
\vspace*{-0.3\baselineskip}
\begin{equation}
	\begin{aligned}
		\label{eq6}
		\text { $Loss$ }&=\sum_{i=1}^n\left(x_i-\hat{x}_i\right)^2
		=\sum_{i=1}^n\left(x_i-\boldsymbol{w}^T A_i\right)^2\\
		&=(\boldsymbol{x}-A \boldsymbol{w})^T(\boldsymbol{x}-A \boldsymbol{w}),
	\end{aligned}
	\vspace*{-0.3\baselineskip}
\end{equation}where $x_i$ and $\hat{x}_i$ represent the actual measured coordinate values and predicted coordinate values, respectively, in the $x$-axis direction for plane 1. The weights of plane 1 are denoted as $\boldsymbol{w}=\left[w_1, w_2, w_0\right]^T$, $A_i=\left[y_i, z_i, 1\right]^T$, $\boldsymbol{A}=\left[A_1, A_2, \ldots, A_n\right]^T$, and $\boldsymbol{x}=\left[x_1, x_2, \ldots, x_n\right]^T$. The $Loss$ function is set to have a partial derivative of zero with respect to $\boldsymbol{w}$.
\vspace*{-0.3\baselineskip}
\begin{equation}
	\label{eq7}
	\frac{\partial L o s s}{\partial \boldsymbol{w}}=2 \boldsymbol{A}^T(\boldsymbol{A} \boldsymbol{w}-\boldsymbol{x})=0.
	\vspace*{-0.3\baselineskip}
\end{equation}

When $\boldsymbol{w}$ satisfies equation (\ref{eq7}), the $Loss$ function achieves its minimum value.
\vspace*{-0.3\baselineskip}
\begin{equation}
	\label{eq8}
	\boldsymbol{w}=\left(\boldsymbol{A}^T \boldsymbol{A}\right)^{-1} \boldsymbol{A}^T \boldsymbol{x}.
	\vspace*{-0.3\baselineskip}
\end{equation}

$\boldsymbol{A}^T \boldsymbol{A}$ is a full rank matrix. $\boldsymbol{w}$ represents the parameters of plane 1, where plane 1 represents the plane that is closest to the sample points in the x-axis direction. Similarly, for plane 2: $y=w_1^{\prime} x+w_2^{\prime} z+w_0^{\prime}$, we can obtain the parameters $\boldsymbol{w}^{\prime}=\left[w_1^{\prime}, w_2^{\prime}, w_0^{\prime}\right]^T$, where plane 2 represents the plane that is closest to the sample points in the $y$-axis direction. By integrating the two plane regression models, the regression line can be obtained.

The right prosthetic hand is taken as the example in this paper, and the participants tend to move toward the right side of the target object when grasping. This indicates that the object to be grasped is located on the left side of the regression line, and the target object is closest to the regression line.

By constructing the separation plane formed by the regression line and the positive direction of the $o-y$ axis as a vector, we can obtain the plane parameters $\boldsymbol{w}_s=\left[w_x, w_y, w_z, w_d\right]^T, $ $\left(w_x<0\right)$. Therefore, when the position of the object is on the left side of the space, it satisfies the following equation:
\vspace*{-0.3\baselineskip}
\begin{equation}
	\label{eq9}
	\boldsymbol{w}_s^T \boldsymbol{p}>0,
	\vspace*{-0.3\baselineskip}
\end{equation}where $\boldsymbol{p}$ 
where $\boldsymbol{p}=\left[x_p, y_p, z_p, 1\right]^T$ is the homogeneous position vector of the object. When there are multiple objects on the left side of the spatial dividing plane, the object closest to the regression line in the left space is determined as the target object. The target object's category $C_{tgt}$, 3D position $P_{tgt}^{3D}$, and size parameters $S_{tgt}^{3D}$ are obtained.

To further evaluate the compatibility between the grasping gestures in the gesture model library and the geometry features of the target object, we perform adaptive point cloud registration between $S_{tgt}^{3D}$ and the corresponding $S_{obj}^{3D}$ in the gesture model library. Initially, the RANSAC (Random Sample Consensus) algorithm \cite{fischler1981random} is utilized for coarse registration, followed by the employment of the ICP (Iterative Closest Point) algorithm \cite{besl1992method} for fine registration. By comparing the differences between $S_{tgt}^{3D}$ and $S_{obj}^{3D}$, the effectiveness of the invoked gesture function $\boldsymbol{F}(\boldsymbol{D})$ is ensured.

\subsubsection{Grasping stage}
For the prosthetic hand system, once the object's category $C_{tgt}$ and size parameters $S_{tgt}^{3D}$ are determined, the corresponding gesture function $\boldsymbol{F}(\boldsymbol{D})$ from the gesture model library is invoked based on the spatial distance $D$ between the object and the hand. This function provides the gesture angles $\boldsymbol{\alpha}=\left[\alpha_p, \alpha_r, \alpha_m, \alpha_i, \alpha_{t b}, \alpha_{t r}\right]$. The system then converts the gesture angles $\boldsymbol{\alpha}$ into pose control signals for the 6 DOFs of the prosthetic hand, which control the flexion and extension motion of the fingers and the rotation of the thumb around its rotation axis. The system continuously adjusts the gesture based on real-time feedback of the hand's pose, enabling the prosthetic hand to perform human-like gestures during the grasping process.

\subsubsection{Grip tightening stage}
The prosthetic hand system integrates metal strain force sensors on the push rod of each DOF to detect the gripping force of the fingers. The effectiveness of the grasp is ensured by setting a grip force threshold. In the grip tightening stage, two situations may occur: a) The preset angles of the gesture function have not been reached, but the grip force threshold has been reached. In this case, the angle of that DOF is locked. b) The preset final angles of the gesture function have been reached, but the preset grip force threshold has not been reached. In this case, the angle of the corresponding DOF will be further adjusted until it reaches the grip force threshold or the angle contraction threshold. In this paper, the grip force threshold is set to 4 N, and the angle contraction threshold is set to 5°.

\section{Experiments}
In this paper, 8 daily objects were chosen as grasp targets, including apple, bottle, bowl, carrot, cup, fork, mouse, and pitcher. Each object requires a different grasping gesture. A total of 7 healthy participants, aged 26±2.3 years, took part in the experiments. Approval of all ethical and experimental procedures and protocols was granted by the Southern University of Science and Technology’s ethical committees. Before the experiments, participants underwent a 30-minute training session to meet the experimental requirements.

\begin{figure}[!t]
	\vspace{-10pt}
	\centering
	\includegraphics[width=0.42\textwidth]{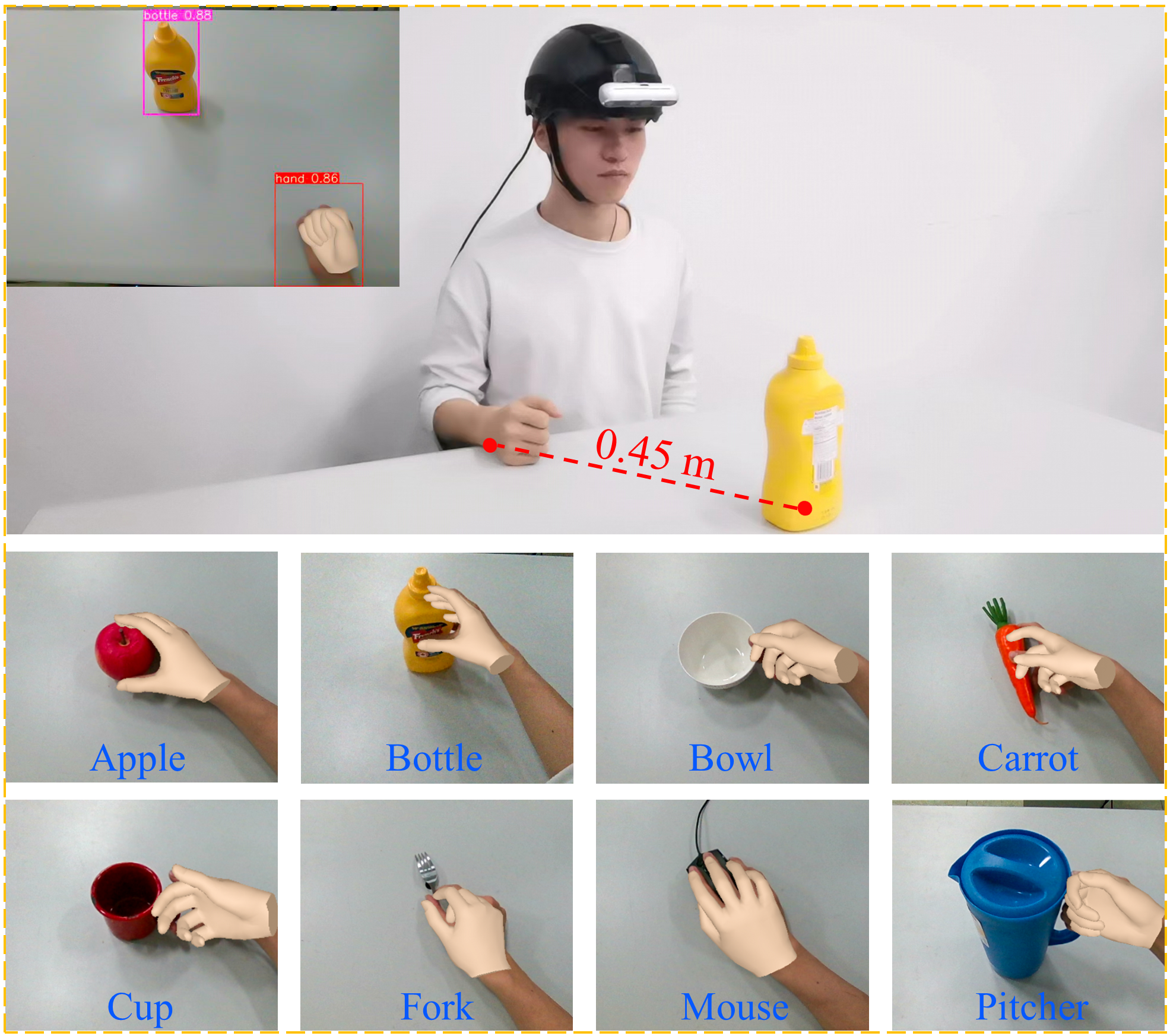}
	\caption{Grasping gesture of participants for constructing the library of the gesture models. The small image in the upper left corner demonstrates that how the camera worn on the head captures object. It demonstrates the natural hand gestures during the process of grasping different objects. Different grasping gestures are adopted to grasp eight objects. The system models the gestures based on the visual information captured by the camera.}
	\label{fig4}
	\vspace{-0.3cm}
\end{figure}

During the grasping experiment, placing the camera in the right front of the participants or mounting it on the head ensures unobstructed capture of finger movements, objects, and gestures. At the initial moment, the participant’s right hand is placed on the table with a semi-closed fist with the object in front of the hand. As presented in Fig. \ref{fig4}, the horizontal distance between the wrist and the center of the object is 0.45 m. It is worth noting that the SG-GM method builds the gesture model with requiring the user to perform only one grasping demonstration. However, to evaluate the duration of human grasping process, each participant was required to perform 20 naturally repeated grasping processes of each object, maintaining the consistency and pace. All participants were required to perform a unified and daily-used grasping gesture in order to minimizing gesture deviation in the experiment. The grasping speed referred to the mirrored natural human speed. The following data were recorded and captured in the experiment. 1) Images of hand grasping objects were captured to build a gesture model library. 2) Time spent and gestures performed during the whole grasping process (from hand starting to moving to holding objects firmly) were recorded to make comparisons with the corresponding data of the prosthetic hand.

\begin{figure}[!t]
	\vspace{-10pt}
	\centering
	\subfloat[]{
		\centering
		\includegraphics[width=0.42\textwidth]{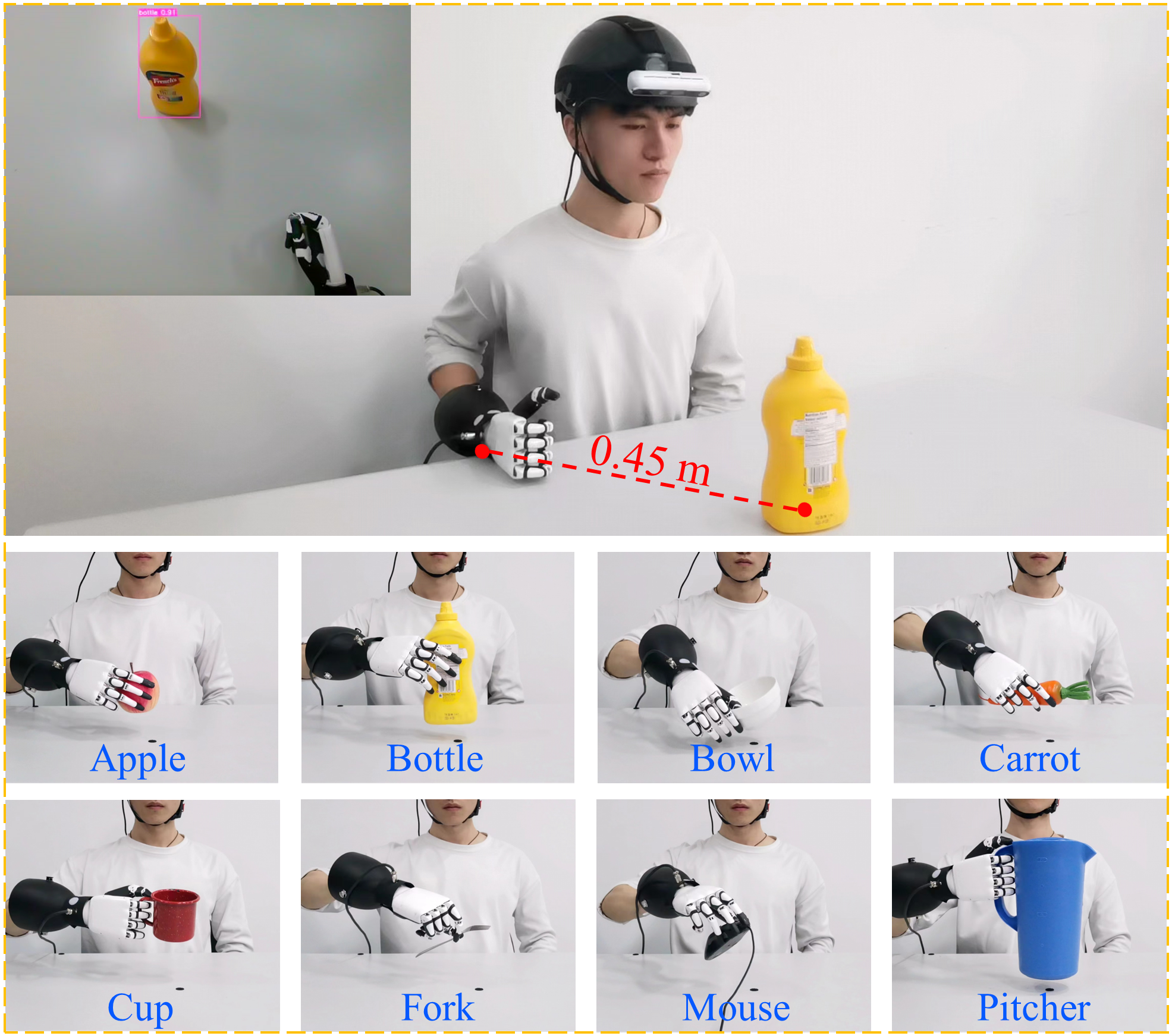}%
		\label{fig5a}}\vspace{-10pt}
	\hfill
	\subfloat[]{
		\centering
		\includegraphics[width=0.42\textwidth]{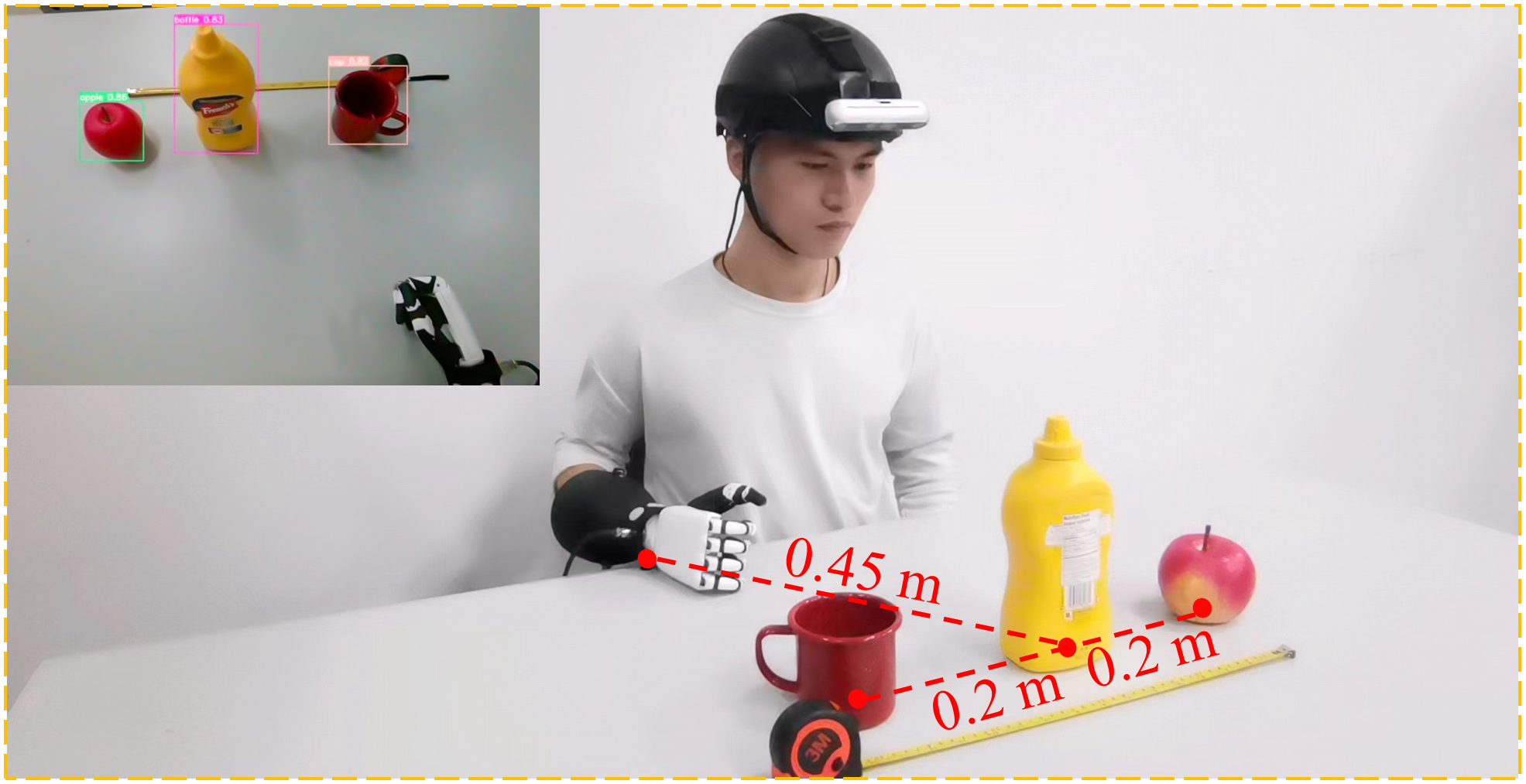}%
		\label{fig5b}}
	\caption{Grasping objects of the prosthetic hand. (a) Single-object scenario and the corresponding prosthetic hand gestures for grasping eight different objects. (b) Multi-object scenario. The initial positions of the prosthetic hands are identical, and there is an equal horizontal spacing between the objects.}
	\label{fig5}
	\vspace{-0.3cm}
\end{figure}

This paper focuses on individuals with upper limb amputations who have intact shoulder and elbow joint functions. The prosthetic hand contains an arm cavity, into which the subjects can insert their residual limb to simulate the condition of patients with missing hands. Two grasping experiments on the prosthetic hand were conducted in this paper: 1) Experiment of grasping a single object (as presented in Fig. \ref{fig5}\subref{fig5a}). This experiment was performed to evaluate the duration and success rate of grasping a single object, as well as the anthropomorphism level of the prosthetic hand gestures; 2) Experiment of grasping multiple objects (as presented in Fig. \ref{fig5}\subref{fig5b}). This experiment was performed to assess the influence of different object spacings in a multi-object environment on accuracy of intent estimation and success rate of grasping. At the beginning of each experiment, the prosthetic hand initially performed a semi-closed fist position and was placed in the same location as in the human hand grasping experiment. During the grasping experiments, the subjects were instructed to turn their heads toward the target objects instead of moving their eyeballs significantly to ensure that the target objects stayed within the camera's field of view. In addition, to ensure consistent grasping gestures, the subjects were asked to maintain consistent arm movements to minimize variations caused by different arm postures. 

\subsection{Experiment of grasping single object}
In each trial of the experiment, only one type of target object appeared. The subject performed 40 grasping actions for each type of object. The system provided real-time feedback on the actual finger pose angles and stored the deviations between the actual and target gestures at different relative distances. During each grasping process, the system automatically recorded the time taken for the grasping action. After the prosthetic hand completed the grasping action, the subject was required to lift the object for approximately 0.1 m. If the object could be lifted steadily and smoothly, it was marked as a successful grasping action (the duration of grasping did not include the time taken to lift the object).

\subsection{Experiment of grasping multiple objects}
Accurate intent estimation is a prerequisite for successful grasping. To evaluate the accuracy of intent estimation and the rate of grasping success at different object spacings, a multi-object grasping experiment was conducted in this paper. The horizontal distance between the centers of objects represents the object spacing. The experiment consisted of 4 sets, with object spacing of 0.3 m, 0.25 m, 0.2 m, and 0.15 m (object volume constrains dictated a minimum spacing of 0.15 m). For each object spacing condition, 5 rounds of experiments were conducted. In each round, three objects were randomly selected from eight options and placed in front of the prosthetic hand. The subject then performed 40 random grasping actions for these three objects. Each experiment recorded the subject's grasping intention, the system's estimated intention, and the result of the grasping action (success or failure). To avoid the bias of human subjective intent from affecting the accuracy of the results, we ensured balance through extensive repeated experiments as described above and randomness by selecting targets in an unpredictable manner.

\subsection{Assessment criteria}
In this paper, the system's performance was evaluated using the following four metrics: 1) Duration of grasping process ($T_p$): This criterion evaluates the time spent for grasping objects $T_p$ of the prosthetic hand. 2) Level of anthropomorphism ($R^2$ and $RMSE$): The similarity between the actual finger angles $\boldsymbol{\alpha}$ during grasping and the hand gesture function $\boldsymbol{F}(\boldsymbol{D})$ is evaluated using the determination coefficient $R^2$ and the root mean square error $RMSE$ \cite{chicco2021coefficient}. 3) Accuracy of intent estimation ($Acc$): This criterion reflects how accurately the system identifies the user's intended object in a multi-object scenario. 4) Success rate of grasping ($Suc$): This criterion indicates the probability of the prosthetic hand successfully grasping an object. The specific calculation formulas for the evaluation metrics are provided in the supplementary materials.

\begin{table*}[]
	\vspace{-15pt}
	\centering
	\caption{Duration of grasping process for the human and prosthetic hand.}
	\label{tab1}
	\scalebox{0.8}{
		\begin{tabular}{cccccccccc}
			\toprule[1 pt]
			& Apple     & Bottle    & Bowl      & Carrot    & Cup       & Fork      & Mouse     & Pitcher & Mean \\ 
			\midrule[1 pt]
			\begin{tabular}[c]{@{}c@{}}Grasping duration\\  for human hand (s)\end{tabular}      & 1.44±0.10 & 1.36±0.13 & 1.35±0.12 & 1.45±0.11 & 1.47±0.11 & 1.48±0.07 & 1.54±0.10 & 1.69±0.10 & 1.47±0.15                      \\ 
			
			\begin{tabular}[c]{@{}c@{}}Grasping duration\\  for prosthetic hand (s)\end{tabular} & 3.24±0.30 & 3.19±0.10 & 2.75±0.17 & 3.24±0.18 & 2.52±0.41 & 3.68±0.18 & 3.01±0.26 & 2.91±0.22	& 3.07±0.41                      \\ 
			\bottomrule[1 pt]
		\end{tabular}
	}
	\vspace{-0.3cm}
\end{table*}

\section{Results and Analysis}
\subsection{Duration of grasping process}
In the single-object environment, the durations of grasping process for both the human hand and the prosthetic hand are presented in Table \ref{tab1}. The average duration of grasping process $T_h$ for the human hand was 1.47 ± 0.15 s. The average duration of grasping process $T_p$ for the prosthetic hand was 3.07 ± 0.41 s. The deviation in the duration of grasping process between the prosthetic hand and the human hand was 1.60 s.
\begin{figure*}[!t]
	\centering
	\includegraphics[width=0.93\textwidth]{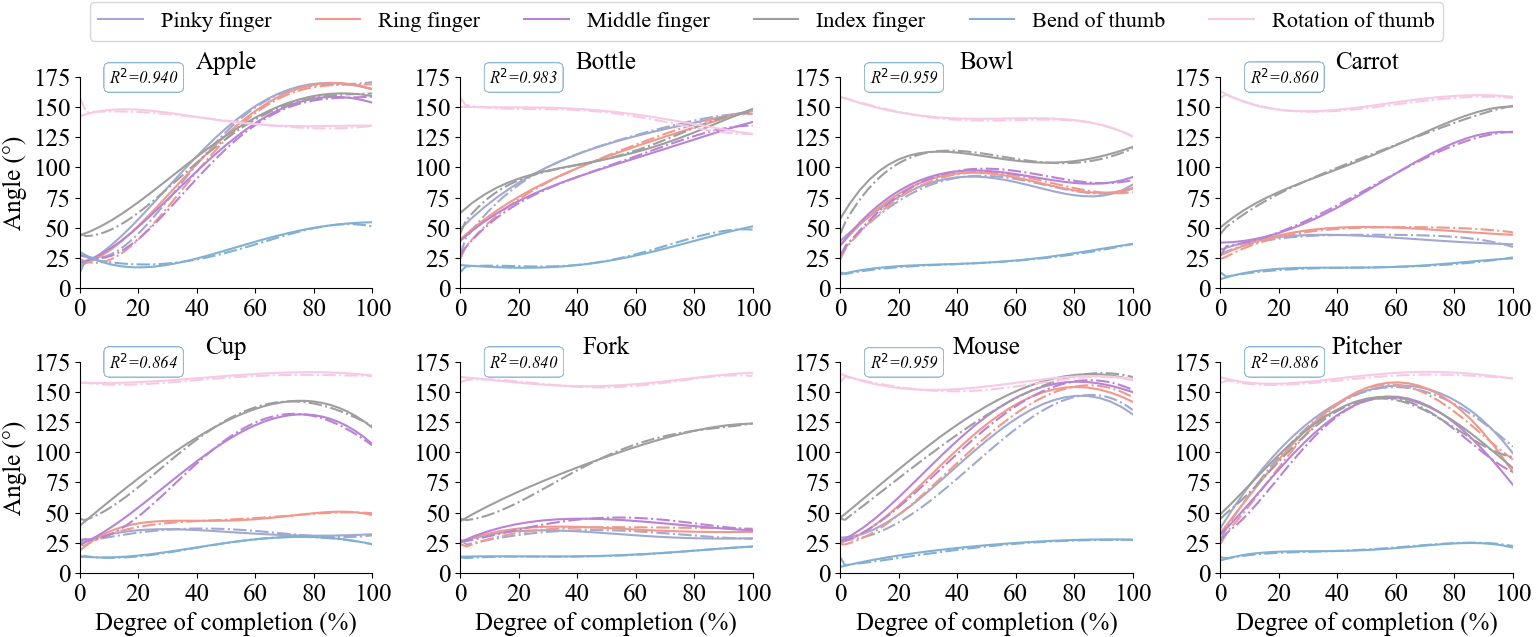}
	\caption{Comparison of grasping gestures between the prosthetic hand and human hand. The solid lines present the motion curves of each DOF of the human hand. The dotted lines present the motion curves of each DOF of the prosthetic hand. The 8 sets of curve graphs correspond to 8 different objects, and each set of curves represents the complete process of gesture transformation during grasping. $R^2$ represents the anthropomorphism index for grasping different objects.}
	\label{fig6}
	\vspace{-0.5cm}
\end{figure*}

\subsection{Level of anthropomorphism}
Due to the influence of different object shapes and sizes and the final grasping gesture, the hand-object distance at the completion of the grasp may vary. To present the changes in hand gestures during the grasping process in a more intuitive and standardized manner, we converted the distance variable into a measure of completion, known as Degree of Completion ($DoC$). The $DoC$ represents the spatial distance progress toward the target throughout the entire grasping stage. The equation is as follows:
\vspace*{-0.3\baselineskip}
\begin{equation}
	\label{eq16}
	DoC=\frac{D_{start}-D_{ongo}}{D_{start}-D_{end}} \times 100\%,
	\vspace*{-0.3\baselineskip}
\end{equation}where $D_{start}$, $D_{ongo}$, and $D_{end}$ represent the hand-object distance at the start, on-going, and end of the grasping stage, respectively.

The comparison of the naturalness of the grasping gestures of the prosthetic hand is presented in Fig. \ref{fig6}. From the figure, it can be observed that the motion trajectories of the prosthetic hand are generally consistent with those of the human hand. The average determination coefficient $R^2$ for the grasping gestures of the 8 objects reached 0.911 with an average $RMSE$ of 2.47\degree. The experimental results indicate that the prosthetic hand is capable of tracking the motion variations of the target gestures, thus performing natural grasping actions effectively.

\subsection{Accuracy of intent estimation}
\begin{figure}[!t]
	\centering
	\includegraphics[width=0.45\textwidth]{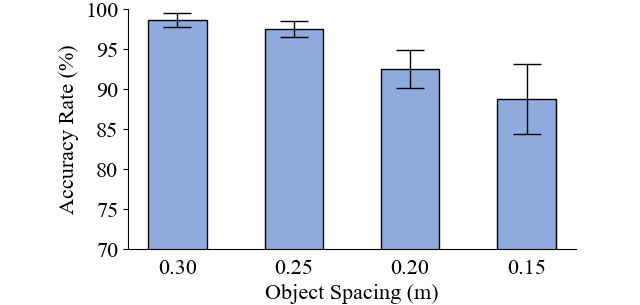}
	\caption{Accuracy of intent estimation with different object spacing.}
	\label{fig7}
	\vspace{-0.3cm}
\end{figure}

The results of intent estimation in a multi-object environment are presented in Fig. \ref{fig7}. The accuracies of intent estimation reached 98.63±0.83$\%$, 97.50±1.00$\%$, 92.50±2.34$\%$, and 88.75±4.39$\%$ at object spacings of 0.3 m, 0.25 m, 0.2 m, and 0.15 m, respectively. As the object spacing decreased, there was a slight decrease in the accuracy of intent estimation. The average accuracy of intent estimation $Acc$ reached 94.35$\%$. These results indicate that the MTR-GIE algorithm is applicable to grasping tasks in multi-object scenarios.

\begin{figure}[!t]
	\centering
	\subfloat[]{\includegraphics[width=0.45\textwidth]{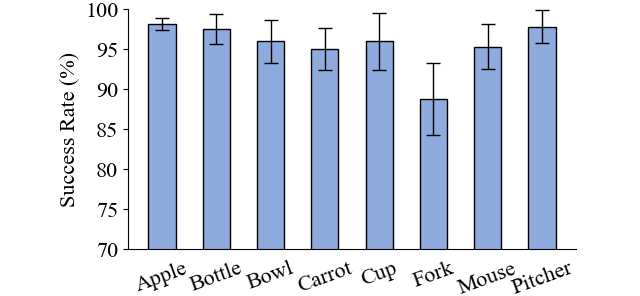}%
		\label{fig8a}}\vspace{-10pt}
	\hfil
	\subfloat[]{\includegraphics[width=0.45\textwidth]{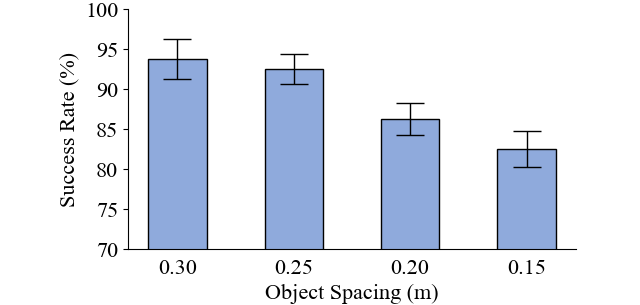}%
		\label{fig8b}}
	\caption{Success rate of grasping objects. (a) Success rate of grasping the 8 objects in the single object experiment. (b) Success rate of grasping multiple objects with different object spacing.}
	\label{fig8}
	\vspace{-0.3cm}
\end{figure}

\subsection{Success rate of grasping}
The results of the grasping success rate in the single-object environment experiment are presented in Fig. \ref{fig8}\subref{fig8a}. The success rates for grasping the apple, bottle, bowl, carrot, cup, fork, mouse, and pitcher reached 98.13±0.75$\%$, 97.50±1.89$\%$, 95.94±2.65$\%$, 95.00±2.67$\%$, 95.94±3.52$\%$, 87.81±4.52$\%$, 95.31±2.81$\%$, and 97.81±2.09$\%$, respectively. The average success rate of grasping was 95.43$\%$. However, the success rate for grasping the fork was the lowest. We speculate that three main factors influence the success rate. The first factor is the complexity of the grasping posture. Different objects have distinct grasping postures. Grasping forks with only two fingers brings the highest difficulty. The second factor is the contact position between the fingers and the object. The more contact points there are, the larger contact area, and the higher the stability of the grasp. The prosthetic hand grasps the fork on both sides of the handle, resulting in only two contact points and a small contact area. The third factor is the size and surface roughness of the object. Generally, smaller objects, complex shapes, and smoother surfaces pose greater difficulty for grasping. These factors also affect the duration of grasping process.

The grasping success rate in the multi-object environment experiment is presented in Fig. \ref{fig8}\subref{fig8b}. The success rate of grasping reaches 93.75±2.55$\%$, 92.50±1.89$\%$, 86.25±2.04$\%$ and 82.50±2.28$\%$ when the object spacing is 0.3 m, 0.25 m, 0.2 m and 0.15 m, respectively. The average grasping success rate $Suc$ for the four object spacings reached 88.75$\%$. The experimental results demonstrate that the success rate in the multi-object grasping experiment is lower than that in the single-object grasping experiment. The surrounding objects cause negative impact on the accuracy of intent estimation. Additionally, the difficulty of grasping objects increases when the object spacing is close.

\section{Discussion}
\subsection{Comparison of three typical control methods}
To comprehensively evaluate the performance of the proposed method in this paper, we have set up three additional typical control methods for comparison:
\begin{itemize}
	\item{Static gesture-based myoelectric control method (Static-MC) \cite{liu2016development, li2021gesture, chen2023review}: We directly define the final static grasping gestures each object. Subjects wear myoelectric sensors to trigger the selection of predefined static gestures. The opening and closing movement of the prosthetic hand are controlled at a constant speed.}
	\item{Vision-based myoelectric control method with the camera on the hand (Hand-VMC) \cite{castro2022continuous, starke2022semi, dovsen2010cognitive, castro2022hybrid, he2020vision}: A camera is mounted below the wrist of the prosthetic hand, and subjects wear an 8-channel myoelectric sensor. Object detection function is implemented in the vision system, along with the intention recognition method proposed by He et al. \cite{he2020vision}. The core of the intention recognition method involves removing distant objects based on depth information from the camera, finding the nearest object to the image center through object detection, and determining the grasping intention by maintaining a certain gaze duration. Upon confirmation of the grasping target, the static grasping gesture is executed through myoelectric triggering for motion control.}
	\item{Vision-based myoelectric control method with the camera on the head (Head-VMC) \cite{shi2023gsi, mouchoux2021artificial, markovic2014stereovision, zandigohar2024multimodal}: This method includes the same object detection function as the Hand-VMC method, and also utilizes myoelectric triggering for motion control of static grasping gestures. The only difference of the two mentioned methods is mounting the camera on the head or hand. The configuration of mounting the camera on the head allows the vision system to capture global environmental features during the grasping process. We adopt the intention recognition method proposed by Mouchoux et al. \cite{mouchoux2021artificial}, which estimates intention by setting spherical regions with the same radius around the centers of each object. When the prosthetic hand moves within the spherical region, the proximity of the hand against all objects is calculated. Therefore, the object with the highest proximity is determined as the grasping target.}
\end{itemize}

In the experiment of grasping single-object, we compared the differences in naturalness and efficiency of grasping motion between these three typical methods and the method proposed in this paper. The experimental setting of the multi-object environment in this study is suitable for comparing methods based on visual control. The Static-MC method lacks visual perception capability, so this method was only compared through single-object grasping experiments. In the experiment of grasping multi-object, we compared the Hand-VMC and Head-VMC methods with the method proposed in this paper to measure the impact of intent estimation methods on system performance in different system configurations.

The results of the comparative experiment are shown in Table \ref{tab2}. Compared with these three methods, the grasping speed of the method proposed in this paper has been increased by more than doubling. The naturalness indicators $R^2$ and $RMSE$ indicate a significant gap between the control method based on static grasping gestures and natural gesture control. However, applying the SG-GM method proposed in this paper to the motion control of multi-degree-of-freedom prosthesis hands can achieve more human-like dynamic grasping gestures. The intent estimation method of Head-VMC performs well when the object spacing is large. However, as the object spacing decreases, the overlapping area between the spheres increases, leading to a significant drop in the fault tolerance of intent estimation. Therefore, this needs to be avoided by unnatural arm movements to prevent intent estimation errors. The intent estimation method of Hand-VMC achieves a comparable high accuracy in intent estimation comparing with the method proposed in this paper. However, the grasping efficiency and naturalness are sacrifices instead. Our MTR-GIE algorithm is capable of quickly and accurately estimating intentions when the prosthetic hand starts to move, which not only saves a considerable amount of time but also guarantees the naturalness of the hand.

\begin{table}[t]
	\centering
	
	\caption{Comparison of experimental results between three typical methods and the method proposed in this paper}
	\label{tab2}
	\begin{threeparttable} 
		\scalebox{0.95}{ 
			\begin{adjustbox}{width=\columnwidth,center} 
				\begin{tabular}{@{}ccccc@{}}
					\toprule[1 pt]
					Method 		& \makecell{Duration of \\ grasping \\ process (s) $\downarrow$} & $R^2$ $\uparrow$ & $RMSE$ (\degree) $\downarrow$  & \makecell{Accuracy of \\ intention \\ estimation (\%) $\uparrow$}  \\ 
					\midrule[1 pt]		
					Static-MC		& 6.14	& $<$ 0 $^*$		& 26.130 	& -  \\ 
					Hand-VMC		& 8.04	& $<$ 0 $^*$		& 27.335 	& 91.63  \\ 
					Head-VMC		& 6.17	& $<$ 0 $^*$		& 24.763 	& 75.69  \\ 
					Ours			& $\boldsymbol{3.07}$	& $\boldsymbol{0.911}$		& $\boldsymbol{2.47}$ 		& $\boldsymbol{94.35}$  \\ 
					\bottomrule[1 pt]
				\end{tabular}
			\end{adjustbox}
		}
		\begin{tablenotes}
			\footnotesize
			\item[1] \parbox{0.94\columnwidth}{$R^2$ $<$ 0 indicates that the actual motion angles of a prosthetic hand are completely deviated from natural hand gestures \cite{chicco2021coefficient}.}
		\end{tablenotes}
	\end{threeparttable}
	\vspace{-0.4cm}
\end{table}

\subsection{Gesture modeling function}
This paper includes a dynamic grasping gesture modeling module, which models the gestures during the process of hand grasping objects via vision sensor. The experimental results demonstrate that the module utilizing the SG-GM method can effectively reproduce the natural gesture changes in the grasping processes. This approach provides a novel solution for enriching grasping patterns. Other than following the predefined and static grasping gestures \cite{castro2022hybrid,zandigohar2024multimodal,shi2020computer, shi2024semi}, the gesture functions obtained through the SG-GM method can represent the angles of each DOF continuously, making them more comprehensive in describing the motion characteristics. To allow upper limb amputees to access grasping gestures, we have provided two methods. 1) Mirror grasping for amputees. Mirror gesture mapping and modeling experiments were conducted to obtain the grasping gestures of the amputee’s healthy hand by mirroring. This method is suitable for single-hand amputees, using the healthy hand’s grasping gestures for the prosthetic hand’s motion control \cite{tully2024validity}. 2) Grasping demonstrations by healthy individuals. Healthy individuals demonstrate grasping gestures to build a gesture model library, which is then applied to amputees by the system.

Although only 8 objects were selected as experimental grasping targets in this paper, the proposed gesture modeling method is universal. It can be expanded to include the gesture model library of objects that can be grasped using a single hand, flexibly accommodating a wide range of grasp patterns to meet personalized demands in different scenarios. The mapping and modeling capabilities of the system greatly reduce the cost of manually defining grasping gestures.

\subsection{Comparative analysis of studies}
The vision sensor, mounted on the head position, provides a global view of the experiments. The vision system observes the position relationship between the prosthetic hand and the object, and provides guidance in improving gestures when the prosthetic hand moves towards the object. Therefore, fast grasping process is implemented. Compared with the method Starke et al. \cite{starke2022semi} proposed that moving the prosthetic hand close to the object and adopting manual control of the grasping gesture with EMG, the method resulted in longer time spent in the grasping processes. With minor differences in experimental settings \cite{starke2022semi}, the method proposed in this paper has significantly shortened the duration of the grasping processes, which is close to that of human hands. A comparison of the grasping duration of related work is provided in the supplementary material, and the evaluation metrics are analyzed.

Amputees generally have high self-esteem and expect to wear a prosthetic hand that moves in an anthropomorphic way rather than a robot \cite{9786528}. Castro et al. \cite{castro2022continuous,shi2020computer,mouchoux2021artificial} set the static grasping gestures and took less consideration of the anthropomorphism of gestures. Starke et al. \cite{starke2022semi} conducted continuous motion control for the prosthetic hand by generating continuous motion trajectories with data gloves. However, the anthropomorphism of the grasping gesture cannot be conducted by the prosthetic hand with only two DOFs. Huang et al. \cite{huang2021development} proposed an EMG-guided continuous control method for the upper-limb prosthesis. Nevertheless, frequent use of EMG signals results in muscle fatigue. In this paper, gesture functions were used as a reference for the motion trajectory of the prosthetic hand, easily achieving process-oriented anthropomorphic gesture control.

\subsection{Limitations}
The method of acquiring gestures with vision sensor only requires low need for hardware settings. However, the contact position and the interaction force between the prosthetic hand and the object are hardly determined. It also brings challenges in performing more stable grasping of the prosthetic hand. The challenge may be solved by introducing a data glove with haptic feedback \cite{sundaram2019learning, liu2019learning}. By deploying the interaction force data on the prosthetic hand with pressure skin, the reliability of performing grasping processes could be improved significantly \cite{cha2022study, zhang2022design, li2023machine}. Additionally, the data glove can help acquire gesture data and avoid losing the blocked gestures by vision sensor.

\section{Conclusion}
The prosthetic hand system with a vision sensor proposed in this paper has the ability to learn anthropomorphic gestures and conduct grasping gesture models for different objects. The gesture control of the prosthetic hand performs anthropomorphic grasping processes and adapts to multi-object scenarios. Additionally, the proposed MTR-GIE algorithm based on motion trajectory regression identified the grasping targets during prosthetic hand movements with high efficiency, and reduced the impact on the duration and anthropomorphism of grasping process. The proposed method provides a fast and anthropomorphic grasping capability for the prosthetic hand to adapt to new objects, offering new approaches and insights for prosthetic hand control exploration. In the future, we will investigate the impact of object pose and grasp location on gestures, exploring the relationship between local features and pose information of objects in gesture modeling. Finally, the method proposed in this paper can be extended to research on wrist and arm pose modeling, which will impact a wider range of application domains.

\vspace{-3pt}
\bibliographystyle{IEEEtran}
\bibliography{manuscript_arXiv.bib}

\onecolumn
\clearpage
\vspace*{0.3cm}
\begin{center}
	\textbf{\huge -- Supplementary Material --}\\
	\vspace{0.5cm}
	\textbf{\huge A Powered Prosthetic Hand with Vision System for Enhancing the Anthropopathic Grasp}
\end{center}
\vspace{1cm}
\begin{table*}[b]
	\centering
	\vspace{-10pt}
	\caption{Comparison of experimental conditions and results for relevant visual control research work}
	\label{tab1}
	\begin{tabular*}{\hsize}{@{}@{\extracolsep{\fill}}cccc@{}}
		\toprule[1 pt]
		\# 		& Study     	& Duration of grasping process (s) & Experimental conditions \\ 
		\midrule[1 pt]		
		\multirow{4}{*}{1}	&\multirow{4}{*}{Starke-2022a\cite{starke2022semi}}	&\multirow{4}{*}{7.9}	&13 daily life objects (selected from the KIT database and YCB dataset). \\
		& & &Hand-object distance of 13 cm. \\
		& & &KIT prosthetic hand with 3 degrees of freedom (including wrist rotation). \\
		& & &Duration of grasping the object. \\ \hline
		
		\multirow{4}{*}{2}	&\multirow{4}{*}{Shi-2020a\cite{shi2020computer}}	&\multirow{4}{*}{6.4}	&20 daily life objects. \\
		& & &Hand-object distance unknown. \\
		& & &HIT-V prosthetic hand with 6 degrees of freedom (4 grasping gestures). \\
		& & &Duration of turning around and grasping the object (excluding the time to select the target). \\	\hline
		
		\multirow{4}{*}{3}	&\multirow{4}{*}{Castro-2022a\cite{castro2022continuous}}	&\multirow{4}{*}{6.8}	&10 daily life objects. \\
		& & &Hand-object distance unknown. \\
		& & &Michelangelo prosthetic hand with 3 degrees of freedom (including wrist rotation). \\
		& & &Duration of grasping, moving, and releasing the object. \\	\hline
		
		\multirow{4}{*}{4}	&\multirow{4}{*}{Markovic-2015a\cite{markovic2015sensor}}	&\multirow{4}{*}{5.9}	&10 daily life objects. \\
		& & &Hand-object distance of 45 cm. \\
		& & &Michelangelo prosthetic hand with 3 degrees of freedom (including wrist rotation). \\
		& & &Duration of grasping the object. \\	\hline
		
		\multirow{4}{*}{5}	&\multirow{4}{*}{Ours}	&\multirow{4}{*}{\textbf{3.07}}	&8 daily life objects. \\
		& & &Hand-object distance of 40 cm. \\
		& & &Inspire prosthetic hand with 6 degrees of freedom (8 grasping gestures). \\
		& & &Duration of grasping the object. \\	
		\bottomrule[1 pt]
	\end{tabular*}
\end{table*}
\begin{multicols}{2}
\subsection{\textbf{Assessment criteria}}
In this paper, the system's performance was evaluated using the following four metrics: duration of grasping process, level of anthropomorphism in hand gestures, accuracy of intent estimation, and success rate of grasping.

\subsubsection{Duration of grasping process}
The duration of grasping process evaluates the time spent for grasping objects $T_p$ of the prosthetic hand, and makes comparison with the duration of grasping process of human $T_h$.

\subsubsection{Level of anthropomorphism}
The similarity between the actual finger angles $\boldsymbol{\alpha}$ during grasping and the hand gesture function $\boldsymbol{F}(\boldsymbol{D})$ is evaluated using the determination coefficient $R^2$ and the root mean square error $RMSE$ \cite{chicco2021coefficient}. The equations for the determination coefficient $R^2_j$ and the $RMSE_j$ for each DOF are listed as following:
\vspace*{-0.3\baselineskip}
\begin{equation}
	\label{eq10}
	R_j^2=1-\frac{\sum_{k=1}^n\left(\alpha_{j, k}-F_j\left(D_k\right)\right)^2}{\sum_{k=1}^n\left(F_j\left(D_k\right)-\overline{F_j}\left(D_k\right)\right)^2},
	\vspace*{-0.3\baselineskip}
\end{equation}
\begin{equation}
	\label{eq11}
	R M S E_{j}=\sqrt{\frac{1}{n} \sum_{k}^{n} \alpha_{j, k}-F_{j}\left(D_{k}\right)} ,
	\vspace*{-0.3\baselineskip}
\end{equation}
where $j \in\{p, r, m, i, t b, t r\}$, $\bar{\alpha}_j$ is the mean value of the hand gesture angles for DOF $j$. To provide an overall assessment of the similarity, the average determination coefficient $R^2$ and the $RMSE$ are calculated by the following equations:
\vspace*{-0.3\baselineskip}
\begin{equation}
	\label{eq12}
	R^2=\frac{1}{6} \sum_j R_j^2,
	\vspace*{-0.3\baselineskip}
\end{equation}
\begin{equation}
	\label{eq13}
	R M S E=\frac{1}{6} \sum_{j} R M S E_{j}.
	\vspace*{-0.3\baselineskip}
\end{equation}

\subsubsection{Accuracy of intent estimation}
This criterion represents the accuracy $Acc$ of the system's intent estimation in multi-object conditions. It is calculated via the following equation:
\vspace*{-0.3\baselineskip}
\begin{equation}
	\label{eq14}
	A c c=\frac{t_{\text{is}}}{t_{\text {total}}} \times 100 \%,
	\vspace*{-0.3\baselineskip}
\end{equation}
where $t_{\text{is}}$ is the number of successful intent estimations under different object spacing, and $t_{\text {total}}$ is the total number of grasping actions performed under that condition.

\subsubsection{Success rate of grasping}
This criterion indicates the probability of the prosthetic hand successfully grasping an object and is calculated as:
\vspace*{-0.3\baselineskip}
\begin{equation}
	\label{eq15}
	S u c=\frac{t_{\text{gs}}}{t_{\text {total }}} \times 100 \%,
	\vspace*{-0.3\baselineskip}
\end{equation}
where $t_{\text{gs}}$ is the number of successful grasping actions under a specified condition. Since successful intent estimation is a prerequisite for successful grasping, in the experiments with multiple objects, if the intent estimation is incorrect, it is directly considered a grasping failure in multi-object environment with incorrect intent estimation.

\subsection{\textbf{Comparison and Analysis of Evaluation Metrics and Benchmark of Related Work}}

Due to the lack of a comprehensive and standardized evaluation system and benchmark for grasping experiments with prosthetic hands, it becomes challenging to establish identical experimental conditions. As a result, researchers often compare experimental results only with a few control methods within the same prosthetic system or set different experimental conditions to validate the system's performance, rather than comparing them with the results of other studies \cite{markovic2015sensor,mouchoux2021artificial,zandigohar2024multimodal,castro2022continuous}. The impact of limited grasping gestures on the success of grasping outcomes is relatively minor. Therefore, these studies focus not on success rates of grasping but often take duration of actions (grasping or performing other actions) as a measure to evaluate control performance \cite{starke2022semi,shi2020computer,castro2022continuous,markovic2015sensor,weiner2022designing,zhong2020reliable}. There are various experimental conditions that affect the duration of grasping process, such as the shape and size of objects, grasping gestures, hand-object distance, types of prosthetic hands, and executed actions. The experimental results of the relevant vision-based grasping control research work are presented in Table \ref{tab1}. The table lists the basic experimental conditions that affect the results, including the types of object, hand-object distance, the types and grasp gestures of the prosthetic hand, and the executed actions. 

Different grasping gestures were employed for different objects in this paper. To validate the effectiveness of diverse grasping gestures, it is necessary to evaluate the grasping success rate as a metric. Additionally, to evaluate the superior performance of the anthropomorphic gesture control method and intent estimation algorithm, we introduced two new evaluation metrics: the level of anthropomorphism in hand gestures and the accuracy of intent estimation. A detailed description of the experimental setup and definition of the evaluation metrics were demonstrated in this paper. Hopefully, a more comprehensive evaluation system and benchmark for prosthetic hand grasping experiments might be established based on our work. This will facilitate future studies to compare their results with the findings presented in this paper, allowing for meaningful comparisons.

\end{multicols}

\end{document}